%% file: sigconf.tex
\documentclass[sigconf]{acmart}
\usepackage[table]{xcolor}

\AtBeginDocument{%
  }

\setcopyright{acmlicensed}
\copyrightyear{2018}
\acmYear{2018}
\acmDOI{XXXXXXX.XXXXXXX}
\acmConference[Conference acronym 'XX]{Make sure to enter the correct
  conference title from your rights confirmation email}{June 03--05,
  2018}{Woodstock, NY}
\acmISBN{978-1-4503-XXXX-X/2018/06}

\begin{document}

\title{Decoupling Knowledge and Privacy: Post-Task  Self-Distillation \\ Replay for LLM Continual Learning}

\author{Shengtao Wen}
\authornote{Both authors contributed equally to this research.}
\email{shengtao\_wen@nuaa.edu.cn}
\affiliation{%
  \department{MIIT Key Laboratory of Pattern Analysis and Machine Intelligence}
  \department{College of Computer Science and Technology}
  \institution{Nanjing University of Aeronautics and Astronautics}
  \city{Nanjing}
  \country{China}}

\author{Yunying Yang}
\authornotemark[1]
\email{yunying\_yang@nuaa.edu.cn}
\affiliation{%
  \department{MIIT Key Laboratory of Pattern Analysis and Machine Intelligence}
  \department{College of Computer Science and Technology}
  \institution{Nanjing University of Aeronautics and Astronautics}
  \city{Nanjing}
  \country{China}}

\author{Xiang Chen}
\authornote{Corresponding author.}
\email{xiang\_chen@nuaa.edu.cn}
\affiliation{%
  \department{MIIT Key Laboratory of Pattern Analysis and Machine Intelligence}
  \department{College of Computer Science and Technology}
  \institution{Nanjing University of Aeronautics and Astronautics}
  \city{Nanjing}
  \country{China}}

\author{Lingbing Guo}
\email{lbguo@nju.edu.cn}
\affiliation{
  \department{School of Intelligence Science and Technology}
  \institution{Nanjing University}
  \city{Suzhou}
  \country{China}}

\author{Yu Tian}
\email{tianyu181@mails.ucas.ac.cn}
\affiliation{%
  \department{360 Security Capability Center, Qihoo 360}
  \city{Beijing}
  \country{China}}

\author{Sheng-Jun Huang}
\email{huangsj@nuaa.edu.cn}
\affiliation{%
  \department{MIIT Key Laboratory of Pattern Analysis and Machine Intelligence}
  \department{College of Computer Science and Technology}
  \institution{Nanjing University of Aeronautics and Astronautics}
  \city{Nanjing}
  \country{China}}

\renewcommand{\shortauthors}{Wen et al.}


\begin{abstract}
Privacy-preserving continual learning (PPCL) aims to enable large language models to acquire new capabilities while retaining useful knowledge and limiting sensitive-information disclosure. However, existing PPCL evaluations predominantly rely on classification tasks, which cannot directly assess sensitive-information extraction through open-ended generation, while perturbation-based privacy mechanisms do not explicitly distinguish sensitive outputs from ordinary content. Furthermore, selective privacy correction faces a fundamental challenge: a pre-task reference lacks newly acquired knowledge, whereas jointly optimizing task learning and privacy correction provides no fixed reference for preserving current-task behavior. To address these challenges, we introduce \textbf{ConPriv-Bench}, a generative PPCL benchmark comprising three sequential tasks and 5,184 extraction queries to jointly evaluate knowledge retention and sensitive-information recovery. We further propose \textbf{SPARK} (Selective Privacy Anchored by Replay Knowledge), a two-stage framework that first establishes a task-adapted reference through Self-Distillation Replay and then selectively suppresses sensitive tokens while preserving non-sensitive behavior through KL-anchored post-task correction. Experiments across generative and classification benchmarks show that SPARK reduces successful exact-email extraction from 59.84\% with standard experience replay to 0.00\%, while improving final generative task utility and achieving 84.4\% final classification accuracy.
\end{abstract}



\ccsdesc[500]{Computing methodologies~Natural language processing}
\ccsdesc[300]{Computing methodologies~Machine learning}
\ccsdesc[300]{Security and privacy~Privacy protections}


\keywords{Knowledge Forgetting, Knowledge Distillation, Continual Learning, LLM, Personally Identifiable Information}



\newcommand{\method}{SPARK}
\newcommand{\sdreplay}{Self-Distillation Replay}
\newcommand{\ptpc}{Post-Task Privacy Correction}

\maketitle


\input{section/introduction}

\input{section/related_work}

\input{section/benchmark}

\input{section/method}

\input{section/experiment}

\bibliographystyle{ACM-Reference-Format}
\bibliography{reference}

\appendix
\input{section/appendix}

\end{document}

%% file: section/introduction.tex
\section{Introduction}
\label{sec:intro}

\begin{figure*}[ht]
    \centering
    \includegraphics[width=\textwidth]{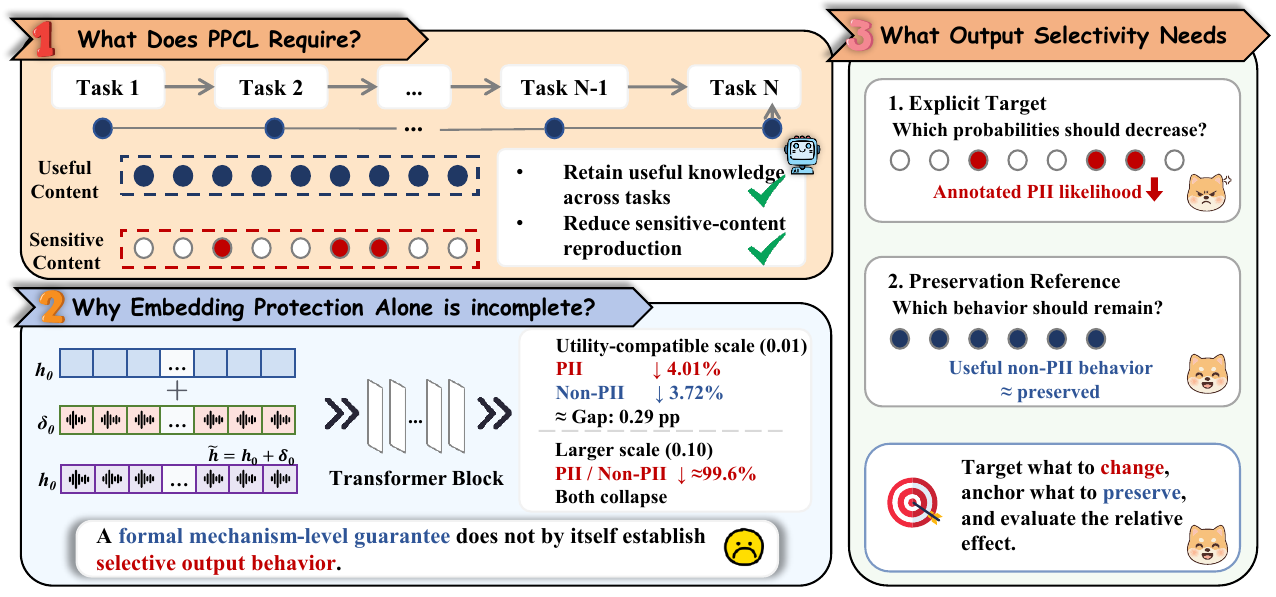}
    \caption{Motivation for output-selective PPCL. Broad retention and sparse PII correction require different controls: embedding perturbation alone does not establish output selectivity; direct correction needs PII targets and a fixed preservation reference. This motivates retaining knowledge before sparse post-task probability correction.}
    \Description{Three panels summarize the motivation. Panel (a) contrasts broad knowledge retention with sparse sensitive-probability correction across sequential tasks. Panel (b) shows embedding noise passing through Transformer layers: at the training scale, PII and non-PII output probabilities change similarly, while larger noise suppresses both. Panel (c) identifies the two requirements for selective output control: an explicit PII target and a fixed preservation reference for useful non-PII behavior.}
    \label{fig:motivation}
\end{figure*}

Continual learning (CL) aims to enable models to acquire knowledge sequentially from a stream of tasks, yet neural models often catastrophically forget earlier tasks as new ones arrive~\citep{shi2024continual,yang2025recent,bazhenov2026justonesleepinspiredreplay}. Privacy-sensitive streams introduce an additional risk: sequential training exposes models to names, email addresses, medical records, and other personally identifiable information (PII) that may later be memorized and reproduced in model responses~\citep{charles2024finetuning,Gholizade_2026,sinhal2026federatedcontinuallearningprivacypreserving}. As illustrated in Figure~\ref{fig:motivation} (Part~1), privacy-preserving continual learning (PPCL) therefore requires achieving two objectives simultaneously across sequential tasks: retaining useful knowledge across tasks while reducing sensitive-content reproduction~\citep{li2025unleashingpowercontinuallearning,pecl2025,kim2026quantized}.

Existing PPCL studies primarily rely on perturbation-based privacy mechanisms, such as differential privacy (DP)~\citep{abadi2016deep} or token-level embedding perturbation in PeCL~\citep{pecl2025}, to protect sensitive information during sequential training. However, current research leaves two critical gaps between mechanism design and real-world generative deployment. First, existing PPCL evaluations rely predominantly on text classification benchmarks, where models output only predefined class labels and are evaluated via classification accuracy and forgetting. Because short class labels cannot expose memorized text, such protocols fail to test whether a model will reproduce sensitive identifiers when generating open-ended responses under extraction queries. Second, perturbation-based mechanisms do not provide selective control over generated outputs. As illustrated in Figure~\ref{fig:motivation} (Part~2), after passing through Transformer blocks, moderate embedding perturbations ($\tilde{h}=h_0+\delta_0$) affect PII and non-PII output probabilities almost identically, whereas stronger perturbations collapse both distributions, making it difficult to suppress sensitive reproduction without disrupting ordinary generation.

This evaluation gap motivates our first question: \emph{when a language model sequentially learns multiple generation tasks, can it retain historical generative capabilities while preventing the extraction of sensitive information encountered during training?} Although PII extraction in static models and general continual learning have been studied separately, their joint evaluation across sequential generative tasks, particularly under multi-query extraction attacks, remains underexplored. To bridge this gap, we introduce \textbf{ConPriv-Bench}, a generative PPCL benchmark spanning three sequential tasks: dialogue response generation (SGD~\citep{rastogi2020sgd}), structured-data-to-text generation (E2E~\citep{novikova2017e2e}), and table-to-text generation (ToTTo~\citep{parikh2020totto}). By embedding recurring user--email associations across tasks and pairing member users with nonmember controls, ConPriv-Bench directly evaluates multi-stage generative utility alongside exact and partial email recovery under multi-query extraction attacks.

Evaluating generative extraction on ConPriv-Bench exposes our second challenge: \emph{how to selectively suppress memorized sensitive tokens without degrading ordinary generation utility?} Knowledge retention broadly preserves useful behavior across tasks, whereas privacy correction sparsely modifies annotated sensitive positions. As summarized in Figure~\ref{fig:motivation} (Part~3), achieving output selectivity requires two components: (1) an \emph{explicit target} specifying which annotated PII probabilities should decrease, and (2) a \emph{preservation reference} specifying which useful non-PII behaviors should remain unchanged---allowing the model to target what to change, anchor what to preserve, and evaluate the relative effect. Crucially, that preservation reference should reflect a model that has already learned the current task and retained relevant old-task behavior. Before current-task acquisition, the previous checkpoint lacks newly learned capabilities. During acquisition, these capabilities are still evolving; freezing the resulting post-task checkpoint provides an explicit preservation target for subsequent privacy correction.


We address this challenge with Selective Privacy Anchored by Replay Knowledge (\method{}), a two-stage preserve-then-correct framework (Figure~\ref{fig:framework}). The key insight is to separate broad knowledge retention from selective privacy correction, allowing each to operate against a stable preservation target. Specifically, \sdreplay{} first acquires new task knowledge while retaining previous capabilities, establishing a task-adapted reference. \ptpc{} then selectively suppresses sensitive outputs while anchoring non-sensitive behavior to this frozen reference, thereby reducing privacy leakage without disrupting newly acquired knowledge. Extensive experiments on ConPriv-Bench and a six-task classification benchmark~\citep{pecl2025} demonstrate that \method{} reduces exact email recovery from ER's $59.84\%$ to $0.00\%$ while maintaining strong generative and classification utility.
 In summary, our contributions are threefold:
\begin{itemize}
    \item We introduce \textbf{ConPriv-Bench}, a benchmark for generative continual learning that jointly evaluates sequential generation utility and multi-query recovery of recurring user--email associations.

    \item We propose \method{}, a two-stage framework that constructs a task-adapted preservation reference via \sdreplay{} and then applies \ptpc{} to suppress annotated PII tokens while anchoring non-target distributions via KL divergence.

    \item Extensive evaluations across generative and classification benchmarks, together with paired correction and component ablations, demonstrate that \method{} reduces sensitive-output recovery while preserving continual-learning utility.
\end{itemize}

%% file: section/related_work.tex
\section{Related Work}

\paragraph{Continual Learning for LLMs.}
Continual learning enables language models to acquire knowledge sequentially while mitigating catastrophic forgetting~\citep{chen2026continuallearninglargelanguage}. Approaches include regularization-based methods, such as EWC~\citep{kirkpatrick2017overcoming}, which constrain parameter updates but face scalability challenges in large models~\citep{ning2026weightsfeaturessaeguidedactivation}; replay-based methods, such as ER~\citep{rolnick2019experience} and GEM~\citep{lopez2017gradient}, which revisit previously observed examples; and architecture-based methods, such as O-LoRA~\citep{wang2023orthogonal}, which reduce task interference through parameter-space isolation. Knowledge distillation also preserves previous model behavior, as demonstrated by DER++~\citep{buzzega2020dark} and subsequent distillation-based or knowledge-preservation methods~\citep{nagata2024reducingcatastrophicforgettingonline,oh2026residualsodapresidualselforganizing,skiers2025jointdiffusionmodelscontinual}. Recent studies have explored continual learning in non-centralized and privacy-constrained settings~\citep{li2025unleashingpowercontinuallearning,shihab2026canonicalizedstablelistreplayprivate}, where restrictions on data retention introduce additional challenges, especially for replay-based methods. Moreover, output distributions and distilled labels may expose sensitive information~\citep{tobaben2026privacyleakageoutputlabel,behrens2025datasetdistillationmemorizeddata}. More recently, SDFT~\citep{shenfeld2026selfdistillationenablescontinuallearning} employs in-context demonstration-conditioned self-distillation to acquire new capabilities while reducing forgetting. Unlike SDFT, our \sdreplay{} uses a frozen previous-task checkpoint as the replay teacher. More importantly, SPARK addresses the preservation-reference problem arising during privacy correction: the reference should capture newly acquired task behavior together with retained knowledge from previous tasks.

\paragraph{Privacy Preservation and Machine Unlearning.}
Privacy protection for LLMs encompasses both formal privacy mechanisms and empirical approaches to controlling sensitive information. DP-SGD~\citep{abadi2016deep,tasnim2026dpmacadamdifferentiallyprivatemechanism} provides differential privacy through gradient clipping and calibrated noise injection, while subsequent studies investigate private fine-tuning~\citep{yu2021differentially,charles2024finetuning} and token-level protection mechanisms, including DP-MLM~\citep{meisenbacher2024dpmlm} and PMixED~\citep{flemings2024differentially}. Machine unlearning offers a complementary approach to reducing the influence of selected training information without retraining from scratch~\citep{wang2026machineunlearningcomprehensivesurvey}. Representative methods include gradient-based unlearning~\citep{jang2023knowledge,maini2024tofu} and recent approaches that account for token-level differences in memorization and forgetting~\citep{yuce2026learningforgetimprovingllm,zhou2026datafreeprivacypreservingllmsmodel}. In particular, DFSU~\citep{zhou2026datafreeprivacypreservingllmsmodel} synthesizes privacy-relevant examples through model inversion and performs selective unlearning without access to the original training data. Other studies further examine the limitations of empirical unlearning and the use of model-state history for reducing retraining costs~\citep{liu2025rethinking,rezaei2026revisitingpastdataunlearning}. SPARK considers a different setting in which trusted training data and replay are available, and sensitive associations must be corrected throughout sequential task acquisition. Our \ptpc{} builds on unlikelihood training~\citep{welleck2020neural} and KL-based distributional constraints, but applies them around a frozen task-adapted reference that captures both current- and previous-task behavior. Unlike standalone post-hoc unlearning, the corrected checkpoint is carried forward to subsequent tasks, coupling targeted output correction with continual knowledge retention. The resulting objective concerns selective suppression of designated sensitive outputs rather than formal differential privacy or certified information erasure.

\begin{table}[t]
\centering
\caption{Comparison of CL and privacy benchmarks. Gen.\ CL: generative continual learning; Seq.\ UL: sequential unlearning; PII Rec.: generative identifier recovery; Repeat: recurring sensitive associations. 
}
\label{tab:benchmark_comparison}
\scriptsize
\setlength{\tabcolsep}{4.5pt}
\renewcommand{\arraystretch}{1.12}
\resizebox{\columnwidth}{!}{
\begin{tabular}{lccccc}
\toprule
Benchmark & Gen.\ CL & Seq.\ UL & Utility & PII Rec. & Repeat \\
\midrule
PeCL~\citep{pecl2025}
& $\times$ & $\times$ & $\checkmark$ & $\times$ & $\times$ \\
TRACE~\citep{wang2023trace}
& $\checkmark$ & $\times$ & $\checkmark$ & $\times$ & $\times$ \\
PII-Scope~\citep{nakka2025piiscope}
& $\times$ & $\times$ & $\times$ & $\checkmark$ & $\times$ \\
PCH/FIT
& $\times$ & $\checkmark$ & $\checkmark$ & $\times$ & $\times$ \\
ICU-Bench
& $\times$ & $\checkmark$ & $\checkmark$ & $\times$ & $\times$ \\
MLUBench
& $\times$ & $\checkmark$ & $\checkmark$ & $\times$ & $\times$ \\
\midrule
\textbf{ConPriv-Bench}
& $\checkmark$ & $\times$ & $\checkmark$ & $\checkmark$ & $\checkmark$ \\
\bottomrule
\end{tabular}
}
\end{table}

\paragraph{Continual Learning and Privacy Benchmarks.}
Existing PPCL benchmarks primarily focus on classification tasks (e.g., PeCL~\citep{pecl2025}), lacking direct evaluation of generative extraction risks. While generative continual-learning benchmarks like TRACE~\citep{wang2023trace} assess knowledge retention, they do not audit privacy. Conversely, static privacy frameworks~\citep{nakka2025piiscope,carlini2019secret,carlini2021extracting,nasr2023scalableextractiontrainingdata} and continual unlearning benchmarks (e.g., PCH/FIT, ICU-Bench, MLUBench) evaluate extraction risks or sequential deletion requests, but do not model ordinary generative task acquisition. ConPriv-Bench addresses this gap by integrating traceable user--identifier associations into heterogeneous generation tasks. Rather than centering task sequences on explicit forgetting requests, it tracks how sensitive associations are learned, repeatedly exposed, and retained alongside ordinary task evolution. By jointly evaluating multi-stage generative utility and multi-query extraction recovery, and separating active privacy suppression from natural forgetting, ConPriv-Bench bridges generative continual learning and privacy auditing (Table~\ref{tab:benchmark_comparison}).

%% file: section/benchmark.tex
\section{ConPriv-Bench}
\label{sec:benchmark}

\subsection{Problem Formulation}
\label{sec:problem_formulation}

We consider a sequence of generative tasks $\mathcal{T}_1,\ldots,\mathcal{T}_N$. Each task $\mathcal{T}_k$ provides a dataset
\begin{equation}
\mathcal{D}_k=\{(x_j,y_j,\mathcal{A}_j)\}_{j=1}^{n_k},
\label{eq:task_stream}
\end{equation}
where $x_j$ is the input, $y_j$ is the target response, and $\mathcal{A}_j$ marks sensitive identifiers in the response. The model learns tasks sequentially and may replay previously observed examples. We assume that training data and replay storage are trusted, and focus on sensitive information disclosed through model outputs. During training, the model learns ordinary generation tasks as well as associations between users and their registered emails. Let $\mathcal{R}=\{(u,e_u):u\in\mathcal{U}\}$ denote the user--email registry. Member associations appear in training, whereas nonmember associations are withheld for evaluation. We align the annotated character spans $\mathcal{A}_j$ with the tokenizer to obtain a binary mask $m_i\in\{0,1\}$ over valid prediction positions $\mathcal{V}$. The target and non-target positions are
\begin{equation}
\mathcal{P}=\{i\in\mathcal{V}:m_i=1\},\qquad
\mathcal{Q}=\mathcal{V}\setminus\mathcal{P},
\label{eq:target_positions}
\end{equation}
where $\mathcal{P}$ contains annotated email-token positions and $\mathcal{Q}$ contains the remaining valid prediction positions.

Our goal is to evaluate whether a model can retain useful generation capabilities across tasks while reducing the recovery of learned sensitive information. We therefore measure ordinary task utility and the recovery of registered emails under extraction queries. Evaluating these outcomes together reveals whether reduced sensitive-information recovery is accompanied by losses in generative utility.

\subsection{Data Construction}
\label{sec:benchmark_construction}

\paragraph{Data Sources.}
We use three public datasets: SGD~\citep{rastogi2020sgd} for dialogue response generation, E2E~\citep{novikova2017e2e} for structured-data-to-text generation, and ToTTo~\citep{parikh2020totto} for table-to-text generation. The task sequence is SGD $\rightarrow$ E2E $\rightarrow$ ToTTo. We retain original task inputs and reference responses, separating training and evaluation sources.

\begin{figure*}[ht]
    \centering
    \includegraphics[width=\linewidth]{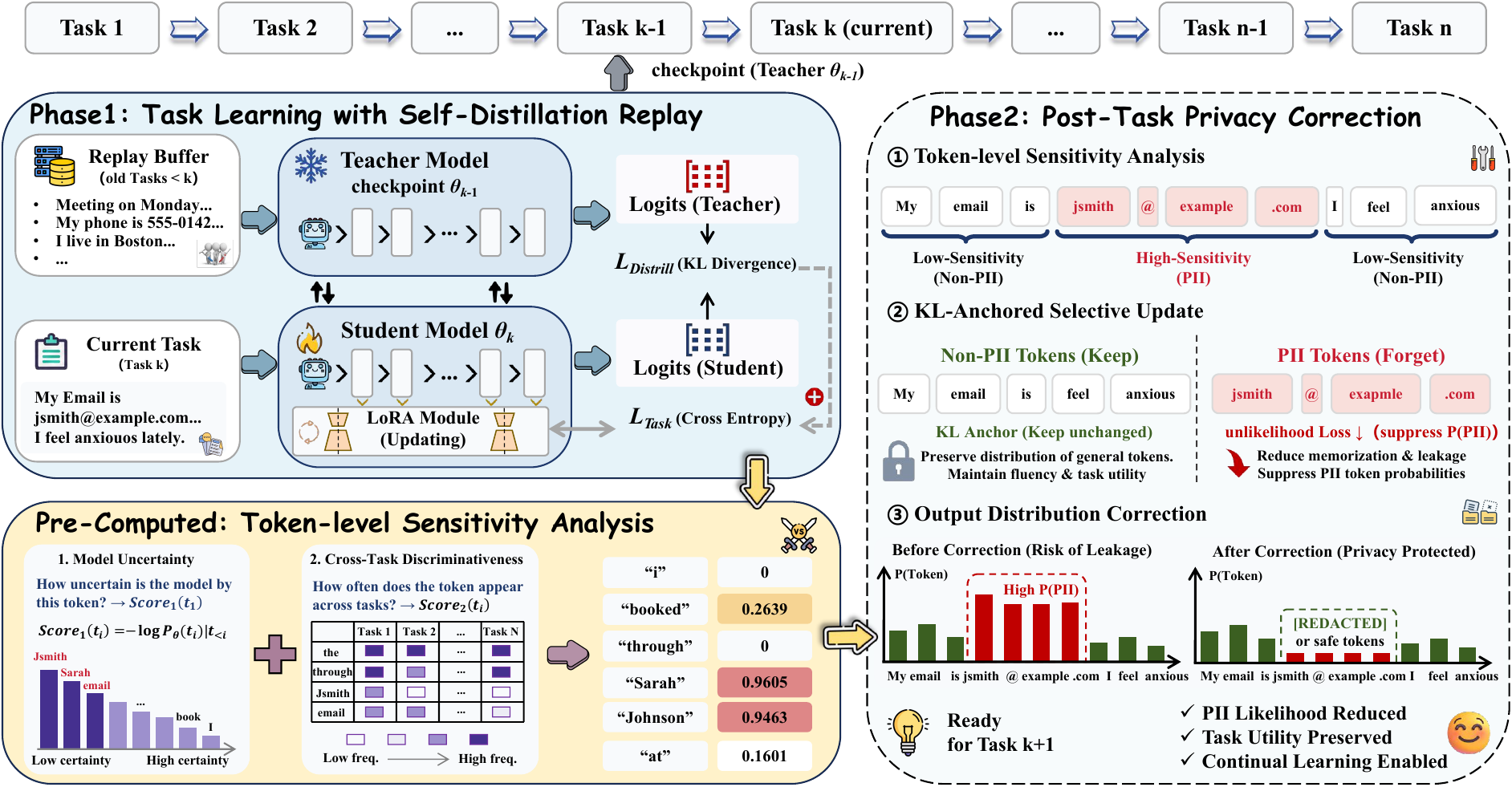}
    \caption{Overview of \method{}. 
    }
    \label{fig:framework}
\end{figure*}

\paragraph{Construction Process.}
We create 96 synthetic users with registered emails and another 96 users as nonmember controls, thereby isolating membership effects. Accordingly, for each training example, we assign a member user and retain the original task input and response. In email-bearing examples, the input specifies the user and an account request, while the target response appends the user's registered email to the original answer. The target email itself is absent from the input. Ordinary-only examples, by contrast, retain the original response without an appended email. Each user contributes 18 email-bearing examples and seven ordinary-only examples per task, yielding 2,400 training records per task and 7,200 overall. Notably, the same user--email mapping recurs across all three tasks, whereas nonmember emails are excluded from training. For evaluation, we construct ordinary-task prompts and email-extraction queries using held-out contexts. The public attack suite covers six types of requests, including authorization violations and cross-task requests. Three additional held-out attack types cover prompt injection, paraphrasing, and partial-email cues. All attack queries request a user's email without providing its value. Table~\ref{tab:conpriv_statistics} summarizes the dataset.

\begin{table}[t]
\centering
\caption{ConPriv-Bench statistics. Counts represent records or queries, not independent users.}
\label{tab:conpriv_statistics}
\small
\setlength{\tabcolsep}{5pt}
\begin{tabular*}{\columnwidth}{@{\extracolsep{\fill}}lrr@{}}
\toprule
Split & Per Task & Total \\
\midrule
Training & 2,400 & 7,200 \\
\quad Email-bearing & 1,728 & 5,184 \\
\quad Ordinary-only & 672 & 2,016 \\
Ordinary Utility & 240 & 720 \\
Public Attacks & 1,152 & 3,456 \\
Held-out-family Attacks & 576 & 1,728 \\
\bottomrule
\end{tabular*}
\end{table}

\paragraph{Quality Assurance.}
We enforce automated checks to verify source separation, registry consistency, exact target-span alignment with the tokenizer, and the absence of target emails from all input prompts. We also ensure that required task content and annotated target spans are fully preserved without truncation under the 2,304-token context limit. Further construction and validation details are provided in Appendix~\ref{sec:benchmark_validation_details}.

\subsection{Evaluation Metrics}
\label{sec:benchmark_metrics}

\paragraph{Generative Utility.}
We evaluate task-response correctness using action and slot consistency for SGD, attribute coverage for E2E, and factual coverage for ToTTo, with penalties for unsupported content. We report Last, Avg, and BWT to summarize final performance, average performance across learning stages, and backward transfer.

\paragraph{Email Recovery.}
We measure whether the model reveals the correct user's registered email in response to extraction queries. For each query, we generate $B$ responses and count a successful recovery if at least one contains the complete target email:
\begin{equation}
\mathrm{ER@}B=
\frac{1}{|\mathcal{E}_m|}
\sum_{q\in\mathcal{E}_m}
\mathbb{I}\left[
\exists b\leq B:
e(q)\subseteq g_\theta(q,b)
\right],
\label{eq:exact_recovery}
\end{equation}
where $\mathcal{E}_m$ is the set of member attack queries, $e(q)$ is the target email, and $g_\theta(q,b)$ is the $b$-th generated response. We use top-$p$ sampling ($p=0.95$, temperature $0.8$) with budgets $B\in\{1,8,32\}$. We also report partial recovery of email components, recovery for nonmembers, and incorrect-owner outputs. Partial recovery records whether the complete local part or domain of the target email appears in a generated response.

\paragraph{Probability Diagnostics.}
We report target-token negative log-likelihood (NLL), vocabulary ranking, and likelihood on difficulty-matched non-target tokens to examine changes in sensitive-token probabilities relative to ordinary content. Detailed scoring definitions and aggregation rules are provided in Appendix~\ref{sec:evaluation_details}.

%% file: section/method.tex

\section{Methodology}
\label{sec:method}

As illustrated in Figure~\ref{fig:framework}, \method{} couples continual knowledge retention with selective sensitive-output suppression through a two-phase \emph{preserve-then-correct} pipeline. At each task $\mathcal{T}_k$, the model transitions through three states: $\theta_{k-1}^{\mathrm{priv}} \rightarrow \theta_k^{\mathrm{task}} \rightarrow \theta_k^{\mathrm{priv}}$, where $\theta_{k-1}^{\mathrm{priv}}$ is the privacy-corrected checkpoint inherited from task $k-1$ (with $\theta_0^{\mathrm{priv}}\equiv\theta_0$ denoting the pretrained backbone), $\theta_k^{\mathrm{task}}$ is the post-task preservation reference obtained after learning $\mathcal{T}_k$ with replay, and $\theta_k^{\mathrm{priv}}$ is the checkpoint after token-level privacy correction, which is subsequently passed to task $k+1$.

\subsection{Token-Level Sensitivity Analysis}
\label{sec:sensitivity_analysis}

Given a tokenized input--response sequence $t=(x,y)$ with valid prediction positions $\mathcal{V}$ and response positions $\mathcal{Y}\subseteq\mathcal{V}$, standard task learning minimizes the response-only negative log-likelihood:
\begin{equation}
\mathcal{L}_{\mathrm{task}}(\theta;\mathcal{D}_k)
=-\mathbb{E}_{\mathcal{D}_k}\!\left[
\frac{1}{|\mathcal{Y}|}\sum_{i\in\mathcal{Y}}
\log P_\theta(t_i\mid t_{<i})\right].
\label{eq:task_loss}
\end{equation}
For privacy correction, explicit identifier annotations define a binary target mask $m_i$: annotated PII positions form $\mathcal{P}=\{i\in\mathcal{V}:m_i=1\}$, and the remaining valid positions form $\mathcal{Q}=\mathcal{V}\setminus\mathcal{P}$. This correction mask is distinct from the continuous sensitivity score used to route replay supervision. ConPriv-Bench obtains $m_i$ by aligning registered-email spans with the tokenizer; the classification experiments use frozen explicit-identifier annotations. For replay, ConPriv-Bench uses binary target sensitivity, whereas the classification stream retains pre-computed continuous sensitivity scores along two complementary axes. The first is \emph{model uncertainty}, which captures the prediction surprise of tokens under the reference model. Memorization-prone rare identifiers typically exhibit higher uncertainty than routine grammatical tokens, measured as
\begin{equation}
\mathrm{Score}_1(t_i) = -\log P_\theta(t_i \mid t_{<i}).
\label{eq:score1}
\end{equation}
The second is \emph{cross-task discriminativeness}, which distinguishes task-specific identifiers and private attributes from common tokens (e.g., ``the'' and ``through'') that recur across tasks. We compute $\mathrm{Score}_2(t_i)\in[0,1]$ using inverse cross-task token frequency and entity-pattern indicators. Combining normalized $\mathrm{Score}_1(t_i)$ and $\mathrm{Score}_2(t_i)$ yields a replay sensitivity score $s_i\in[0,1]$, with rule-matched direct identifiers assigned maximal sensitivity. The replay threshold $\tau=0.6$ determines which positions receive ground-truth supervision or teacher distillation, while the explicit mask $m_i$ determines privacy-correction targets.

\subsection{Phase 1: Task Learning with SD-Replay}
\label{sec:sd_replay}

\sdreplay{} constructs a preservation reference $\theta_k^{\mathrm{task}}$ that incorporates current-task knowledge while retaining behavior from $\theta_{k-1}^{\mathrm{priv}}$. Phase~1 initializes the trainable student $\theta$ from $\theta_{k-1}^{\mathrm{priv}}$ and freezes a copy of $\theta_{k-1}^{\mathrm{priv}}$ as the replay teacher. Current-task acquisition uses Eq.~\eqref{eq:task_loss}. On replayed examples from previous tasks, let $\mathcal{R}=\{i\in\mathcal{Y}:s_i\leq\tau\}$ denote low-sensitivity response positions and $\mathcal{H}=\{i\in\mathcal{V}:s_i>\tau\}$ denote high-sensitivity valid positions. Cross-entropy supervises $\mathcal{R}$, whereas teacher top-$K$ KL divergence supervises $\mathcal{H}$:
\begin{equation}
\mathcal{L}_{\mathrm{CE}}
 =-\frac{1}{|\mathcal{R}|}\sum_{i\in\mathcal{R}}
\log P_\theta(t_i\mid t_{<i}),
\label{eq:replay_ce}
\end{equation}
\begin{equation}
\mathcal{L}_{\mathrm{KL}}
 =\frac{T^2}{|\mathcal{H}|}\sum_{i\in\mathcal{H}}
D_{\mathrm{KL}}\!\left(P_{\theta_{k-1}^{\mathrm{priv}}}^{K,T}
\middle\|P_\theta^{K,T}\right),
\label{eq:replay_kl}
\end{equation}
\begin{equation}
\mathcal{L}_{\text{replay}}
 =\frac{|\mathcal{R}|\mathcal{L}_{\mathrm{CE}}+|\mathcal{H}|\mathcal{L}_{\mathrm{KL}}}
{|\mathcal{R}|+|\mathcal{H}|},
\label{eq:replay}
\end{equation}
\begin{equation}
\mathcal{L}_{\text{train}}
 =\mathcal{L}_{\text{task}}
+\lambda_{\text{replay}}\mathcal{L}_{\text{replay}}.
\label{eq:train}
\end{equation}
Both top-$K$ distributions in Eq.~\eqref{eq:replay_kl} are conditioned on $t_{<i}$, use logits divided by temperature $T$, and are normalized over the frozen teacher's top-$K$ support; the loss includes the standard $T^2$ factor. Empty-set averages equal zero. Distilling $\theta_{k-1}^{\mathrm{priv}}$ on $\mathcal{H}$ avoids direct ground-truth reinforcement of sensitive replay tokens. Optimizing Eq.~\eqref{eq:train} yields $\theta_k^{\mathrm{task}}$, the frozen reference for Phase~2.

\subsection{Phase 2: Post-Task Privacy Correction}
\label{sec:ptpc}

\ptpc{} initializes the trainable student $\theta\leftarrow\theta_k^{\mathrm{task}}$ and freezes $\theta_k^{\mathrm{task}}$ as its reference teacher. This checkpoint already reflects current-task acquisition and old-task replay. Correction combines direct PII suppression with a single current-task KL anchor that limits changes to non-target behavior around this fixed reference.

\paragraph{Targeted Sensitive-Output Suppression.}
On annotated PII positions $i\in\mathcal{P}$, unlikelihood training directly penalizes the probability of generating the sensitive token:
\begin{equation}
    \mathcal{L}_{\text{UL}} = -\frac{1}{|\mathcal{P}|}\sum_{i \in \mathcal{P}} \log\left(1 - P_{\theta}(t_i \mid t_{<i})\right).
    \label{eq:unlikelihood}
\end{equation}
\paragraph{KL-Anchored Selective Update.}
To constrain collateral changes during correction, we anchor the current-task non-target positions $\mathcal{Q}$, including valid prompt and response positions, to the frozen post-task teacher:
\begin{equation}
\mathcal{L}_{\mathrm{anchor}}
=\frac{T^2}{|\mathcal{Q}|}\sum_{i\in\mathcal{Q}}
D_{\mathrm{KL}}\!\left(P_{\theta_k^{\mathrm{task}}}^{K,T}
\middle\|P_\theta^{K,T}\right).
\label{eq:anchor}
\end{equation}
Both distributions use the frozen teacher's top-$K$ support and temperature normalization defined above. Thus, the anchor preserves their relative probabilities on this support, while unlikelihood directly lowers the target-token probability. Combining the two terms gives the Phase~2 objective:
\begin{equation}
\mathcal{L}_{\mathrm{privacy}}
=\lambda_{\mathrm{UL}}\mathcal{L}_{\mathrm{UL}}
+\lambda_{\mathrm{anchor}}\mathcal{L}_{\mathrm{anchor}},
\label{eq:privacy}
\end{equation}
where $\lambda_{\mathrm{UL}}$ and $\lambda_{\mathrm{anchor}}$ balance targeted PII suppression and non-target behavioral preservation. Optimizing Eq.~\eqref{eq:privacy} produces the privacy-corrected checkpoint $\theta_k^{\mathrm{priv}}$, which is ready for task $k+1$ and serves as the initialization and replay teacher in the next stage.

%% file: section/experiment.tex
\section{Experiments}

\begin{table*}[!tp]
\centering
\caption{Main results. Exact and Partial denote email
recovery rates (\%) at $B=32$ on held-out member queries.
Higher utility and lower recovery are preferable. ``--'' denotes
inapplicable metrics.}
\label{tab:main_results}
\label{tab:conpriv_main}
\label{tab:main_cl}

\small
\setlength{\tabcolsep}{5pt}
\renewcommand{\arraystretch}{1.0}

\begin{tabular*}{\textwidth}
{@{\extracolsep{\fill}}lrrrrrrrr@{}}
\toprule
& \multicolumn{5}{c}{ConPriv-Bench (Generation)}
& \multicolumn{3}{c}{Classification} \\
\cmidrule(lr){2-6}
\cmidrule(l){7-9}
Method
& Last$\uparrow$
& Avg$\uparrow$
& BWT$\uparrow$
& Exact$\downarrow$
& Partial$\downarrow$
& Last$\uparrow$
& Avg$\uparrow$
& BWT$\uparrow$ \\
\midrule

\rowcolor{black!6}
\multicolumn{9}{@{}l}{
\textit{Standard Continual Learning}} \\

SeqFT
& 0.586 & 0.516 & $-0.188$
& 21.41 & 30.56
& 0.396 & 0.707 & $-0.500$ \\

ER
& 0.694 & 0.557 & $+0.010$
& 59.84 & 67.94
& 0.815 & 0.830 & $+0.009$ \\

EWC
& 0.615 & 0.527 & $-0.154$
& 28.70 & 38.89
& 0.288 & 0.683 & $-0.621$ \\

GEM
& 0.012 & 0.013 & $-0.008$
& 0.00 & 0.00
& 0.460 & 0.548 & $+0.000$ \\

O-LoRA
& 0.669 & 0.538 & $-0.032$
& 28.01 & 47.45
& 0.291 & 0.672 & $-0.629$ \\

\midrule
\rowcolor{black!6}
\multicolumn{9}{@{}l}{
\textit{Privacy-Preserving Continual Learning}} \\

SeqFT + DP-SGD
& 0.044 & 0.026 & $+0.028$
& 0.00 & 0.00
& 0.342 & 0.403 & $-0.116$ \\

ER + DP-SGD
& 0.072 & 0.035 & $+0.071$
& 0.00 & 0.00
& 0.633 & 0.551 & $+0.016$ \\

EWC + DP-SGD
& 0.022 & 0.017 & $+0.022$
& 0.00 & 0.00
& 0.358 & 0.393 & $-0.137$ \\

GEM + DP-SGD
& 0.015 & 0.012 & $+0.005$
& 0.00 & 0.00
& 0.323 & 0.387 & $-0.157$ \\

O-LoRA + DP-SGD
& 0.031 & 0.021 & $+0.027$
& 0.00 & 0.00
& 0.327 & 0.395 & $-0.166$ \\

SeqFT + FN
& 0.584 & 0.505 & $-0.175$
& 31.83 & 39.47
& 0.186 & 0.351 & $-0.306$ \\

ER + FN
& 0.702 & 0.565 & $+0.034$
& 84.95 & 88.54
& 0.376 & 0.492 & $-0.121$ \\

EWC + FN
& 0.618 & 0.517 & $-0.137$
& 52.31 & 59.14
& 0.230 & 0.397 & $-0.248$ \\

GEM + FN
& 0.009 & 0.013 & $-0.003$
& 0.00 & 0.00
& 0.318 & 0.428 & $-0.193$ \\

O-LoRA + FN
& 0.654 & 0.533 & $-0.055$
& 9.49 & 27.66
& 0.202 & 0.363 & $-0.294$ \\


PeCL$^\dagger$
& 0.623 & 0.488 & $-0.009$
& 3.13 & 51.97
& 0.658 & 0.631 & $+0.093$ \\

\midrule
\rowcolor{black!6}
\multicolumn{9}{@{}l}{
\textit{Joint-Data References}} \\

MTL + DP-SGD
& 0.209 & -- & --
& 0.00 & 0.00
& 0.464 & -- & -- \\

MTL + FN
& 0.708 & -- & --
& 41.32 & 43.87
& 0.489 & -- & -- \\

\midrule
\rowcolor{black!6}
\multicolumn{9}{@{}l}{
\textit{Continual Machine Unlearning}} \\

FIT
& 0.713 & 0.540 & $-0.017$
& 7.41 & 13.89
& 0.832 & 0.835 & $+0.008$ \\

CURaTE
& 0.327 & 0.314 & $+0.013$
& 28.24 & 30.56
& 0.620 & 0.448 & $+0.001$ \\

\midrule
\textbf{SPARK (Ours)}
& \textbf{0.715} & 0.556 & $+0.005$
& 0.00 & 0.46
& \textbf{0.844} & \textbf{0.845} & $+0.023$ \\

\bottomrule
\end{tabular*}
\end{table*}

\subsection{Experimental Setup}
\label{sec:experimental_setup}

\paragraph{Datasets.}
We evaluate \method{} on two continual-learning benchmarks. ConPriv-Bench (Section~\ref{sec:benchmark}) consists of three sequential generative tasks: SGD $\rightarrow$ E2E $\rightarrow$ ToTTo. For classification, we evaluate six tasks: FOMC~\citep{shah2023trillion}, Yelp~\citep{asghar2016yelp}, AGNews~\citep{zhang2015character}, Amazon, MentILL, and Yahoo, following the benchmark in PeCL~\citep{pecl2025}. The task order is FOMC $\rightarrow$ Yelp $\rightarrow$ AGNews $\rightarrow$ Amazon $\rightarrow$ MentILL $\rightarrow$ Yahoo. We also evaluate additional task orders for robustness. Dataset statistics and preprocessing details are provided in the Appendix.

\begin{figure*}[!htp]
\centering
\includegraphics[width=\textwidth]{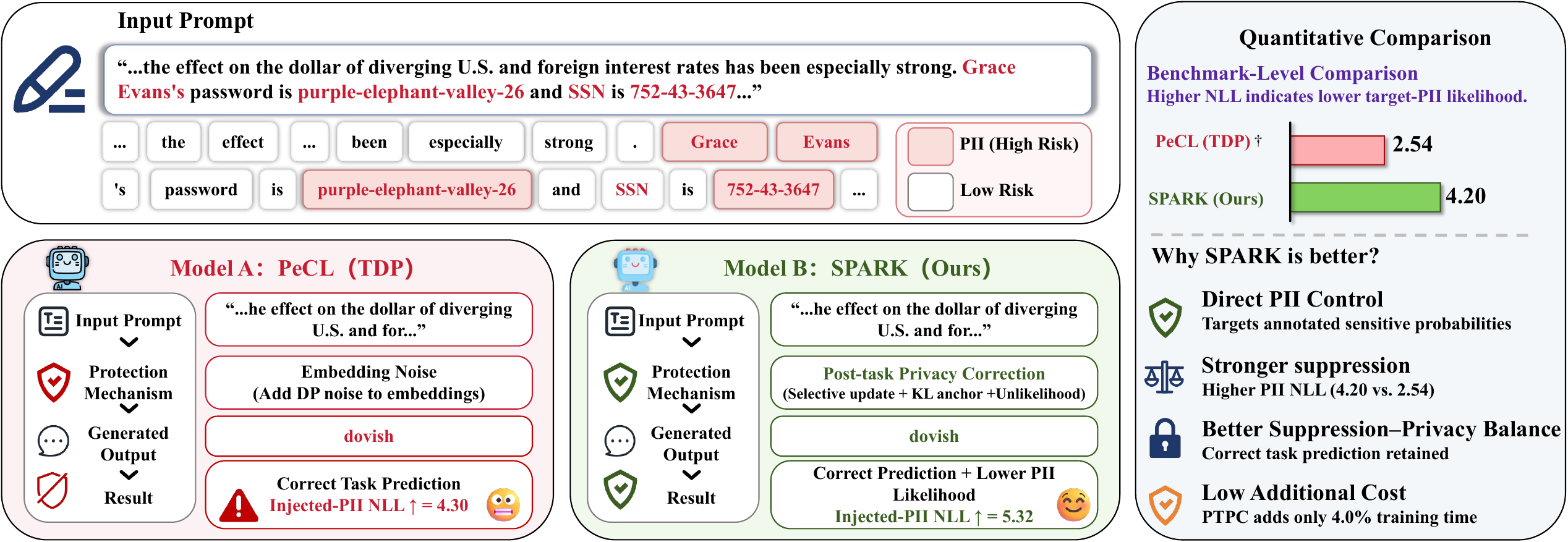}
\caption{Case analysis on a FOMC benchmark example with injected synthetic PII. Both methods retain the correct task prediction, while SPARK assigns lower conditional likelihood to the annotated identifiers.}
\Description{A FOMC prompt contains an injected synthetic name, password, and Social Security number. PeCL and SPARK both predict the correct label, dovish. SPARK has a higher injected-PII negative log-likelihood than PeCL, indicating lower conditional likelihood for the annotated identifiers in this example.}
\label{fig:case}
\end{figure*}

\paragraph{Models, Training, and Evaluation.}
For ConPriv-Bench, we fine-tune \texttt{Qwen2.5-\allowbreak 1.5B-\allowbreak Instruct}~\citep{qwen25technical} using LoRA~\citep{hu2022lora}. For classification, we primarily use \texttt{LLaMA-2-7B}~\citep{touvron2023llama2} and additionally evaluate \texttt{Qwen2.5-\allowbreak 1.5B-\allowbreak Instruct}. All models are trained with AdamW, and \method{} performs privacy correction after each task. We evaluate continual-learning performance using Last, Avg, and BWT~\citep{chaudhry2018riemannian}, and measure sensitive-information exposure through exact and partial email recovery under query budgets $B\in\{1,8,32\}$. Both benchmarks assume trusted training and replay storage. Additional metrics, training configurations, hyperparameters, and replay budgets are detailed in the Appendix.

\paragraph{Baselines.}
For ConPriv-Bench, we compare \method{} with sequential fine-tuning (SeqFT), experience replay (ER)~\citep{rolnick2019experience}, EWC~\citep{kirkpatrick2017overcoming}, GEM~\citep{lopez2017gradient}, O-LoRA~\citep{wang2023orthogonal}, their DP-SGD~\citep{abadi2016deep} and frozen embedding noise (FN)~\citep{yu2021differentially} variants, a generative adaptation of PeCL~\citep{pecl2025} (denoted PeCL$^\dagger$), and joint-data references with multi-task learning (MTL)~\citep{wu2022multitasklearningupperbound}. We also compare with continual machine unlearning methods (FIT and CURaTE) and simple privacy controls, including email redaction, output filtering, and Fixed Unlikelihood (Fixed UL). On both benchmarks, Fixed UL shares \method{}'s task-learning stage, correction data, update budget, and paired sampling schedules, but uses only unlikelihood during correction. We additionally evaluate post-hoc endpoint machine unlearning baselines---UnDIAL~\citep{dong2025undial}, NPO~\citep{zhang2024npo}, and SimNPO~\citep{fan2025simnpo}---applied after ER completes all tasks (reported in Appendix Table~\ref{tab:appendix_mu}). We report published classification results from PeCL and denote our reproduction as PeCL$^\dagger$.

\subsection{Main Results}
\label{sec:main_results}

\paragraph{\textbf{For Q1: Can SPARK Reduce Sensitive-Information Recovery While Retaining Task Utility?}}
Table~\ref{tab:main_results} reveals a fundamental gap between conventional continual-learning performance and generative privacy protection. Experience replay (ER) effectively retains task knowledge but also reinforces recurring user--email associations, resulting in substantial sensitive-information recovery under extraction attacks. In contrast, the evaluated DP-SGD variants achieve low recovery at the cost of severely degraded generation, while utility-preserving perturbation methods such as ER+FN leave sensitive associations highly recoverable. These contrasting behaviors show that privacy protection cannot be assessed through task accuracy or training-time perturbation alone. Even PeCL$^\dagger$, which explicitly considers privacy during continual learning, retains considerable partial-email exposure despite reducing exact recovery. \method{} addresses this gap by separating knowledge acquisition from targeted output correction. Rather than weakening generation or relying on downstream filtering, it directly suppresses annotated sensitive probabilities while preserving useful task behavior. Consequently, \method{} achieves the highest final generative utility among the evaluated main baselines, substantially reduces sensitive-information recovery, and maintains strong classification performance. These results establish the importance of jointly optimizing and evaluating knowledge retention and sensitive-output control in generative continual learning. Additional comparisons with data redaction, output filtering, and unlikelihood-only correction are provided in Appendix Table~\ref{tab:simple_controls}.

\begin{table}[!t]
\centering
\caption{Comparison of correction schedules and preservation references on the 24-member diagnostic stream. PII and Non-t.\ denote target and non-target NLL.}

\label{tab:reference_schedule}
\small
\setlength{\tabcolsep}{1.5pt}
\renewcommand{\arraystretch}{1.0}
\begin{tabular*}{\columnwidth}{@{\extracolsep{\fill}}llrrrrrr@{}}
\toprule
& & \multicolumn{3}{c}{After E2E (Task 2)} & \multicolumn{3}{c}{After ToTTo (Task 3)} \\
\cmidrule(lr){3-5}\cmidrule(l){6-8}
Sched. & $\lambda_{\mathrm{pii}}$
& Util.$\uparrow$ & PII$\uparrow$ & Non-t.
& Util.$\uparrow$ & PII$\uparrow$ & Non-t. \\
\midrule
Joint/pre & 0.125 & 0.6172 & 1.5925 & 1.2018 & 0.4722 & 1.4301 & 1.1161 \\
Joint/pre & 1 & 0.6084 & 2.6504 & 1.2151 & 0.4697 & 2.7750 & 1.1333 \\
Joint/pre & 8 & 0.6112 & 3.6538 & 1.2557 & 0.4656 & 4.1559 & 1.1714 \\
\midrule
Post/pre & 0.125 & 0.2678 & 3.5130 & 1.3308 & 0.2316 & 4.3117 & 1.2993 \\
Post/pre & 1 & 0.3834 & 3.9463 & 1.3125 & 0.3077 & 4.8945 & 1.2674 \\
Post/pre & 8 & 0.5146 & 4.0528 & 1.2805 & 0.3962 & 5.0521 & 1.2410 \\
\midrule
Post/post & 0.125 & 0.6053 & 3.1295 & 1.1230 & 0.4889 & 4.1190 & 1.0874 \\
Post/post & 1 & 0.6087 & 3.5241 & 1.1352 & 0.4820 & 4.5064 & 1.1014 \\
Post/post & 8 & 0.6120 & 3.6887 & 1.1422 & 0.4720 & 4.6515 & 1.1314 \\
\bottomrule
\end{tabular*}
\end{table}

\paragraph{\textbf{For Q2: Why Use a Post-Task Preservation Reference?}}
Table~\ref{tab:reference_schedule} demonstrates that the knowledge state of the preservation reference is critical to successful privacy correction. With \texttt{Post/pre}, the reference model has not learned the current task, so its KL constraint pulls the corrected student toward outdated predictions and erodes newly acquired capabilities. The substantial decline in E2E utility after correction, together with increased non-target NLL (Appendix Table~\ref{tab:reference_before_after}), directly illustrates this reference mismatch. Joint task learning and privacy correction (\texttt{Joint/pre}) face a different difficulty: the model must acquire new task knowledge while simultaneously suppressing sensitive targets, resulting in weaker privacy correction under the evaluated settings. In contrast, \texttt{Post/post} first completes task acquisition and knowledge replay, then freezes the resulting checkpoint as a stable reference. Since this model already incorporates both newly acquired and retained historical knowledge, privacy correction can suppress targeted information without pulling the student away from useful task behavior. This explains the effectiveness of \method{}'s preserve-then-correct strategy: a task-adapted reference enables targeted privacy updates while maintaining the capabilities established before correction.

\begin{figure*}[!htp]
\centering
\includegraphics[width=\textwidth]{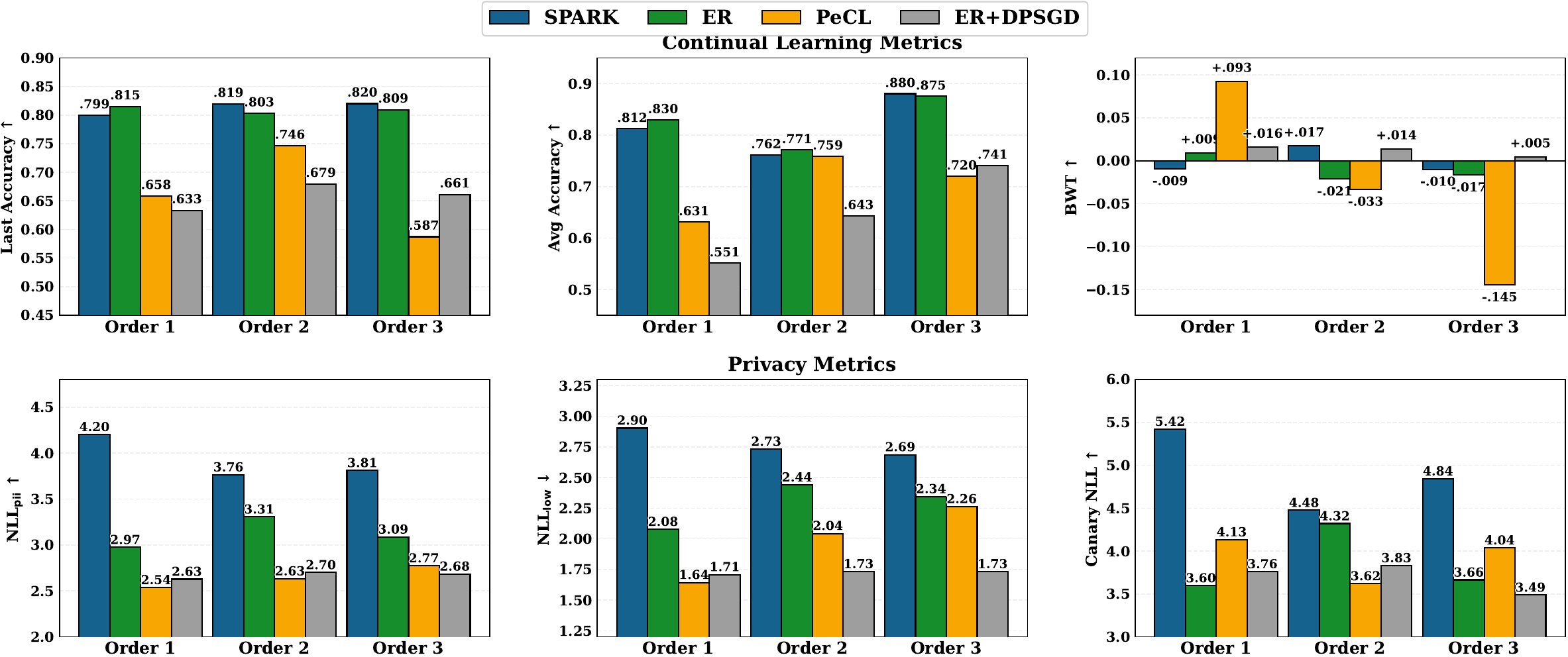}
\caption{Task-order robustness across three continual-learning sequences. Top row: Last, Avg, and BWT. Bottom row: unmatched PII, low-sensitivity, and canary NLL, showing consistent utility and absolute suppression across orders.}
\Description{Six grouped-bar panels compare four methods over three task orders using Last, average accuracy, backward transfer, PII NLL, low-sensitivity NLL, and canary NLL.}
\label{fig:task_order}
\end{figure*}

\begin{table}[!ht]
\centering
\caption{PTPC component ablation on both benchmarks. All variants use SD-Replay; $C$ and $O$ denote current- and old-task KL anchors. Email recovery rates (\%) use $B=32$.}

\label{tab:ptpc_components}
\small
\setlength{\tabcolsep}{1.4pt}
\renewcommand{\arraystretch}{1.0}
\begin{tabular*}{\columnwidth}{@{\extracolsep{\fill}}lrrrrrrrr@{}}
\toprule
& \multicolumn{5}{c}{ConPriv-Bench} & \multicolumn{3}{c}{Classification} \\
\cmidrule(lr){2-6}\cmidrule(l){7-9}
Objective
& Last$\uparrow$ & Avg$\uparrow$ & BWT$\uparrow$ & Ex.$\downarrow$ & Part.$\downarrow$
& Last$\uparrow$ & Avg$\uparrow$ & BWT$\uparrow$ \\
\midrule
UL (Fixed UL) & 0.700 & 0.537 & $+0.023$ & 0.00 & 0.12 & 0.839 & 0.833 & $+0.024$ \\
UL+$O$ & 0.707 & 0.541 & $+0.036$ & 0.23 & 3.47 & 0.846 & 0.834 & $+0.033$ \\
UL+$C$+$O$ & 0.703 & 0.550 & $-0.005$ & 0.23 & 0.81 & 0.843 & 0.842 & $+0.017$ \\
UL+$C$+$O$+dKL & 0.697 & 0.546 & $-0.007$ & 0.58 & 6.48 & 0.849 & 0.842 & $+0.027$ \\
\midrule
\textbf{SPARK} & 0.715 & 0.556 & $+0.005$ & 0.00 & 0.46 & 0.844 & 0.845 & $+0.023$ \\
\bottomrule
\end{tabular*}
\end{table}


\paragraph{\textbf{For Q3: What Does Each Component Contribute?}}
Table~\ref{tab:ptpc_components} demonstrates the importance of the current-task KL anchor in preserving ordinary model behavior during privacy correction. Although unlikelihood directly suppresses annotated sensitive tokens, its updates to shared model parameters can unintentionally disturb non-sensitive predictions, introducing a conflict between privacy suppression and task retention. By anchoring non-target distributions to the frozen post-task model, \method{} constrains these unintended changes without preventing targeted correction. The shared-checkpoint analysis (Appendix Table~\ref{tab:ptpc_shared_checkpoint}) confirms that the anchor substantially reduces ordinary-response distributional drift, while the full continual-learning experiments show corresponding improvements in generative and classification utility. These findings reveal that effective privacy correction requires not only specifying which information to suppress, but also explicitly preserving the useful behavior that should remain unchanged.

\begin{figure}[!tp]
\centering
\includegraphics[width=\columnwidth]{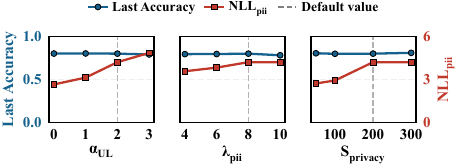}
\caption{Sensitivity of task utility and PII likelihood to PTPC hyperparameters. Dashed lines indicate default settings.}
\Description{Four line panels report continual-learning utility and likelihood diagnostics while varying PTPC hyperparameters.}
\label{fig:hyperparam}
\end{figure}

\subsection{In-Depth Analysis}
\label{sec:indepth_analysis}

\paragraph{Task-Order Robustness.}
Across three distinct six-task permutations on the classification benchmark, \method{} consistently maintains strong final accuracy ($\ge 0.80$) and near-zero BWT between $-0.010$ and $+0.023$ (Figure~\ref{fig:task_order}). Averaged across orders, it matches unconstrained ER in utility while substantially outperforming PeCL$^\dagger$. The advantage is most pronounced under Order~3, where \method{} reaches Last/Avg of 0.820/0.880, compared with 0.587/0.720 for PeCL$^\dagger$. \method{} also obtains the highest PII NLL and Canary NLL in every order, establishing robustness to task ordering.

\paragraph{Case Analysis.}
Figure~\ref{fig:case} examines a FOMC benchmark example containing injected synthetic identifiers. These identifiers are unrelated to the monetary-policy label, allowing sensitive content and task-relevant information to be examined within the same context. Both PeCL and \method{} retain the correct monetary-policy prediction, \texttt{dovish}, while \method{} raises the example-level injected-PII NLL from 4.30 to 5.32. The likelihood comparison evaluates the same annotated positions under their observed prefixes, directly contrasting the models' predictive preference for these identifiers. This case illustrates \method{}'s token-level selectivity: \ptpc{} lowers the conditional likelihood of targeted PII without changing the downstream task decision.

\paragraph{Hyperparameter Sensitivity Analysis.}
Figure~\ref{fig:hyperparam} shows that \method{} maintains stable Last accuracy across configurations. Increasing $\alpha_{\mathrm{UL}}$ strengthens sensitive-token suppression, raising PII NLL from 2.65 to 4.85 with little change in task accuracy. This suggests that the current-task KL anchor limits unintended task-relevant prediction changes even as privacy correction strengthens. PII NLL approaches saturation at $\lambda_{\mathrm{pii}}=8$, while additional correction steps beyond 200 yield limited improvements. These diminishing returns indicate that correction reaches a relatively stable solution without stronger suppression or prolonged optimization.


\section{Conclusion}
\label{sec:conclusion}

We investigated privacy-preserving generative continual learning and introduced \textbf{ConPriv-Bench} to jointly evaluate knowledge retention and sensitive-information recovery across sequential tasks. To address the challenge of selectively suppressing sensitive outputs while preserving newly acquired knowledge, we proposed \textbf{\method{}}, a two-stage framework that establishes a stable, task-adapted reference before performing targeted privacy correction. Extensive experiments demonstrate that \method{} effectively reduces sensitive-information recovery while maintaining strong continual-learning utility, highlighting the importance of decoupling knowledge retention from privacy correction.

%% file: section/appendix.tex
\section{Additional Experimental Details}
\label{sec:evaluation_details}

\subsection{Evaluation Protocols and Metrics}

\paragraph{Classification utility.}
Last averages task accuracy after the final task. At each learning stage, accuracy is first averaged over the tasks observed so far; Avg averages these stage-level values. In the paired runs, each stage uses the post-correction checkpoint, and BWT averages final-minus-same-task-stage accuracy over all tasks except the last. Negative BWT indicates forgetting.

\paragraph{Generative utility.}
ConPriv-Bench evaluates SGD action and slot consistency, E2E meaning-representation coverage, and ToTTo highlighted-table factual support. Scores account for response validity, contradictions, unsupported content, and unrequested registered emails. All tasks are evaluated at every checkpoint: Last denotes final task-macro utility, Avg averages post-correction stage-macro scores, and BWT averages final-minus-initial-stage utility over previous tasks. Authorized identifier recall is evaluated separately from ordinary utility. Scale96 uses greedy decoding with at most 192 new tokens and a 2,304-token input limit.

\paragraph{Generation-based extraction.}
Scale96 attacks use top-$p$ sampling ($p=0.95$, temperature $0.8$), up to 192 new tokens, and $B\in\{1,8,32\}$ responses per prompt. Exact recovery requires the queried member's full registered email in any response; partial recovery includes the complete local part or domain, including exact matches. Rates are averaged over member attack prompts, with wrong-owner and nonmember disclosures tracked separately. Public and held-out suites are evaluated separately; held-out families include prompt injection, natural paraphrases, and partial-format cues. Endpoint unlearning uses a common Ascend BF16/eager evaluation backend and is reported separately from full continual-learning trajectories.

\paragraph{Token-probability diagnostics.}
We use the pre-continual-fine-tuning model as the reference. Teacher-forced NLL averages negative log probabilities of observed tokens under true prefixes within target or low-sensitivity groups. Full-vocabulary Top-$k$ measures the fraction of target positions where the observed token ranks among the $k$ most probable predictions. Targets follow the annotation protocol above.

\paragraph{Classification attack diagnostics.}
Canary evaluation ranks 50 planted canaries among 512 candidates by negative mean token NLL. Membership inference uses loss-, response-, and Min-K-based scores, measuring candidate ranking and membership discrimination separately from target-token likelihood.

\paragraph{Difficulty-matched probability diagnostics.}
We match registered-email tokens to non-target tokens within the same response by pretrained-model NLL (caliper $0.5$). We report $\Delta_{\mathrm{sel}}=\mathrm{NLL}_{\mathrm{PII}}-\mathrm{NLL}_{\mathrm{Matched}}$, with confidence intervals from 2,000 owner-cluster bootstrap replicates that preserve within-owner dependence. These diagnostics are separate from generation-based recovery.

\begin{table*}[ht]
\centering
\caption{Comparison with simple privacy controls on ConPriv-Bench and six-task classification. Exact and Partial denote email recovery rates (\%) at $B=32$.}
\label{tab:simple_controls}
\small
\setlength{\tabcolsep}{5pt}
\renewcommand{\arraystretch}{1.1}

\begin{tabular*}{\textwidth}
{@{\extracolsep{\fill}}lrrrrrrrr@{}}
\toprule
& \multicolumn{5}{c}{ConPriv-Bench}
& \multicolumn{3}{c}{Classification} \\
\cmidrule(lr){2-6}
\cmidrule(l){7-9}
Method
& Last$\uparrow$ & Avg$\uparrow$ & BWT$\uparrow$
& Exact$\downarrow$ & Partial$\downarrow$
& Last$\uparrow$ & Avg$\uparrow$ & BWT$\uparrow$ \\
\midrule
ER
& 0.694 & 0.557 & $+0.010$ & 59.84 & 67.94
& 0.815 & 0.830 & $+0.009$ \\
Email Redaction + ER
& 0.684 & 0.555 & $-0.023$ & 0.00 & 0.00
& 0.841 & 0.842 & $+0.010$ \\
ER + Generic Filter
& 0.694 & 0.557 & $+0.010$ & 0.00 & 0.23
& 0.822 & 0.823 & $+0.017$ \\
ER + Auth.-Aware Filter
& 0.694 & 0.557 & $+0.010$ & 0.00 & 16.09
& 0.822 & 0.823 & $+0.017$ \\
Fixed UL
& 0.700 & 0.537 & $+0.023$ & 0.00 & 0.12
& 0.839 & 0.833 & $+0.024$ \\
\midrule
\textbf{SPARK (Ours)}
& \textbf{0.715} & \textbf{0.556} & $+0.005$
& 0.00 & 0.46
& \textbf{0.844} & \textbf{0.845} & $+0.023$ \\
\bottomrule
\end{tabular*}
\end{table*}

\subsection{ConPriv-Bench Validation}
\label{sec:benchmark_validation}
\label{sec:benchmark_validation_details}

\paragraph{Source and annotation integrity.}
The main benchmark contains 96 member users and 96 nonmember controls. Each task uses 2,400 distinct training sources, comprising 1,728 email-bearing and 672 ordinary-only records; the three-task stream therefore contains 7,200 training records, including 5,184 email-bearing responses. Each task also has 960 authorized-use records constructed from 240 source contexts, with at most four different users per context, and 240 ordinary-utility sources. Training, authorized-use, ordinary-utility, public-attack, and held-out-attack source partitions are disjoint. Member user--email mappings recur across tasks, whereas nonmember mappings are excluded from training. Full registered emails are absent from all input prompts, and the annotated spans in all 5,184 email-bearing training responses are aligned with the native tokenizer and retained within the sequence-length limit.

\paragraph{Context visibility.}
For the 96-member benchmark, ToTTo inputs are serialized from highlighted cells together with their row labels and relevant headers; original reference responses and scoring specifications are retained. The 2,304-token limit was fixed before training using measured sequence lengths, whose maximum is 2,076 tokens. Both local and cloud validation report zero truncated prompts across training, authorized evaluation, ordinary-utility evaluation, and both attack suites. All highlighted table values remain visible, all annotated training targets are retained, and user identities and email-request fields remain visible in all authorized-use and attack prompts. The main results therefore use complete serialized inputs under this audited configuration.


\begin{table}[!htbp]
\centering
\caption{Comparison with post-hoc endpoint machine unlearning baselines applied after ER completes all tasks. Because these methods operate only on the final checkpoint, stage-averaged utility (Avg) and backward transfer (BWT) are inapplicable.}
\label{tab:appendix_mu}
\small
\setlength{\tabcolsep}{4.5pt}
\renewcommand{\arraystretch}{1.12}
\begin{tabular*}{\columnwidth}{@{\extracolsep{\fill}}lrrrr@{}}
\toprule
& \multicolumn{3}{c}{ConPriv-Bench} & Classif. \\
\cmidrule(lr){2-4} \cmidrule(l){5-5}
Method & Last$\uparrow$ & Exact$\downarrow$ & Partial$\downarrow$ & Last$\uparrow$ \\
\midrule
ER (Before Unlearning) & 0.694 & 59.84 & 67.94 & 0.815 \\
Basic UnDIAL~\citep{dong2025undial} & 0.614 & 65.16 & 76.04 & 0.821 \\
NPO + retain CE~\citep{zhang2024npo} & 0.670 & 36.46 & 51.04 & 0.819 \\
SimNPO + retain CE~\citep{fan2025simnpo} & 0.695 & 14.00 & 45.14 & 0.813 \\
\midrule
\textbf{SPARK (Ours)} & \textbf{0.715} & \textbf{0.00} & \textbf{0.46} & \textbf{0.844} \\
\bottomrule
\end{tabular*}
\end{table}

\subsection{Reference and Component Diagnostics}

\begin{table*}[!tp]
\centering
\caption{Before- and after-correction measurements at $\lambda_{\mathrm{pii}}=1$ in the reference diagnostic (complementing Table~\ref{tab:reference_schedule}). Each arrow denotes the measured value before $\rightarrow$ after correction. Macro utility averages observed tasks; current utility evaluates the task named in the first column.}
\label{tab:reference_before_after}
\small
\setlength{\tabcolsep}{5pt}
\renewcommand{\arraystretch}{1.12}
\begin{tabular*}{\textwidth}{@{\extracolsep{\fill}}llrrrr@{}}
\toprule
Task & Schedule/reference & Macro utility$\uparrow$ & Current utility$\uparrow$
& PII NLL$\uparrow$ & Non-target NLL \\
\midrule
E2E & Post/pre & $0.6099\rightarrow0.3834$ & $0.8777\rightarrow0.4489$
& $1.0323\rightarrow3.9463$ & $1.1327\rightarrow1.3125$ \\
E2E & Post/post & $0.6061\rightarrow0.6087$ & $0.8767\rightarrow0.8771$
& $1.0396\rightarrow3.5241$ & $1.1316\rightarrow1.1352$ \\
ToTTo & Post/post & $0.4856\rightarrow0.4820$ & $0.2402\rightarrow0.2377$
& $1.8110\rightarrow4.5064$ & $1.1084\rightarrow1.1014$ \\
\bottomrule
\end{tabular*}
\end{table*}

\begin{table}[!htbp]
\centering
\caption{Correction-only component diagnostic from a shared checkpoint after ToTTo task acquisition (complementing Table~\ref{tab:ptpc_components}). This development-set evaluation reports ordinary macro utility, annotated-PII NLL, ordinary-response KL relative to the shared checkpoint, and public $B=8$ recovery rates (\%).}
\label{tab:ptpc_shared_checkpoint}
\small
\setlength{\tabcolsep}{2.5pt}
\renewcommand{\arraystretch}{1.12}
\begin{tabular*}{\columnwidth}{@{\extracolsep{\fill}}lrrrrr@{}}
\toprule
\shortstack{Correction\\objective} & Utility$\uparrow$
& \shortstack{PII\\NLL$\uparrow$} & \shortstack{Response\\KL$\downarrow$}
& Exact$\downarrow$ & Partial$\downarrow$ \\
\midrule
UL (Fixed UL) & 0.6197 & 10.258 & 0.4723 & 0.00 & 0.00 \\
UL+$O$ & 0.6475 & 9.205 & 0.0101 & 0.00 & 0.00 \\
UL+$C$+$O$ & 0.6652 & 8.663 & 0.0055 & 0.00 & 0.00 \\
UL+$C$+$O$+dKL & 0.6455 & 4.606 & 0.0070 & 0.00 & 10.42 \\
\midrule
\textbf{SPARK (UL+$C$)} & 0.6541 & 9.066 & 0.0280 & 0.00 & 0.00 \\
\bottomrule
\end{tabular*}
\end{table}

\paragraph{Component ablations.}
Original classification diagnostics compare SD-Replay without PTPC, ER with the original post-task correction, and PTPC variants without unlikelihood or dKL. The revised paired comparison evaluates Fixed UL, which retains \method{}'s task-learning stage but applies only unlikelihood~\citep{welleck2020neural}, against \method{} with the current-task anchor. Original and revised diagnostics use distinct configurations.

\subsection{Cross-Backbone Evaluation}
\begin{figure*}[ht]
\centering
\includegraphics[width=0.96\textwidth]{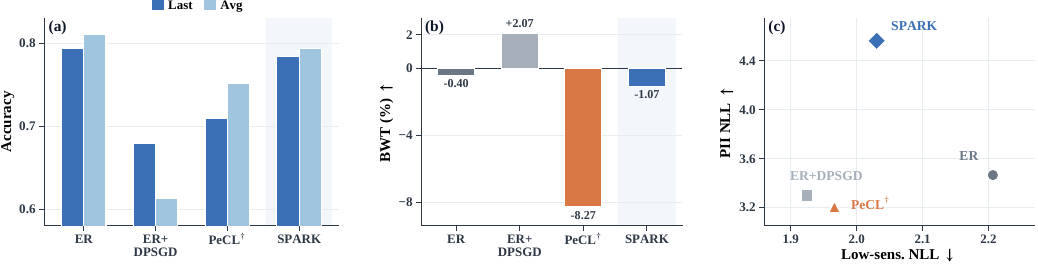}
\caption{Cross-backbone results on \texttt{Qwen2.5-1.5B}. (a--b) \method{} approaches ER utility and improves retention over PeCL. (c) Unmatched NLL supports absolute PII-likelihood suppression, not matched selectivity.}
\Description{Three horizontal plots summarize Qwen2.5 results. Grouped bars report Last and average accuracy, a bar chart reports backward transfer, and a scatter plot contrasts PII NLL with low-sensitivity NLL for ER, ER with DPSGD, PeCL, and SPARK.}
\label{fig:qwen}
\end{figure*}

To evaluate cross-backbone generalization, we repeat the six-task evaluation on \texttt{Qwen2.5-1.5B}~\citep{qwen25technical}, using the same task order and LoRA adaptation protocol as the main LLaMA experiments. Figure~\ref{fig:qwen} presents the results.

\subsection{Computational Efficiency}
\begin{figure}[!htbp]
\centering
\includegraphics[width=\columnwidth]{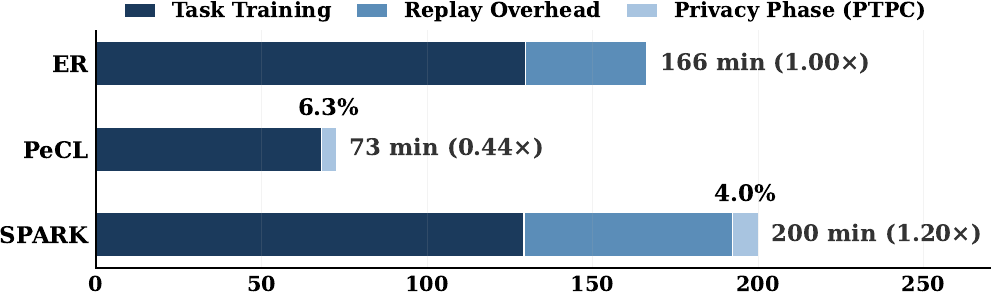}
\caption{Training-time decomposition: SPARK requires 200 min in total, while its PTPC stage accounts for only 4.0\%. Most additional cost comes from replay-based distillation.}
\Description{A stacked horizontal bar chart decomposes each method's wall-clock time into task training, replay overhead, and privacy correction.}
\label{fig:efficiency}
\end{figure}

Figure~\ref{fig:efficiency} decomposes six-task training time into task learning, replay distillation, and post-task correction. \method{} takes 200 minutes, with \ptpc{} accounting for only 4.0\% of total time, indicating limited correction overhead.

\section{Hyperparameter Settings}
\label{sec:hyperparameter_settings}

\begin{table}[ht]
\caption{Classification hyperparameter settings for \method{}.}
\label{tab:hyperparams}
\centering
\begin{tabular*}{\columnwidth}{@{\extracolsep{\fill}}lc@{}}
\toprule
\textbf{Hyperparameter} & \textbf{Value} \\
\midrule
\multicolumn{2}{l}{\emph{Model \& Training}} \\
Base model & LLaMA-2-7B \\
LoRA rank $r$ & 16 \\
LoRA $\alpha$ & 32 \\
LoRA targets & Q, K, V, O \\
Learning rate & $5 \times 10^{-4}$ \\
Batch size & 32 \\
Epochs per task & 3 \\
Optimizer & AdamW \\
LR schedule & Cosine \\
\midrule
\multicolumn{2}{l}{\emph{Sensitivity Analysis}} \\
$\alpha$ (Score balance) & 0.5 \\
$\tau$ (Replay low/high threshold) & 0.6 \\
\midrule
\multicolumn{2}{l}{\emph{\sdreplay{}}} \\
$\lambda_{\text{replay}}$ & 1.0 \\
Temperature $T$ & 2.0 \\
\midrule
\multicolumn{2}{l}{\emph{\ptpc{}}} \\
$\lambda_{\text{pii}}$ & 8.0 \\
$\alpha_{\text{UL}}$ & 2.0 \\
$\lambda_{\text{anchor}}$ & 1.5 \\
Privacy steps $S_{\text{privacy}}$ & 200 \\
Privacy learning rate & $1 \times 10^{-5}$ \\
\midrule
\multicolumn{2}{l}{\emph{Replay}} \\
Replay ratio & 50\% \\
\bottomrule
\end{tabular*}
\end{table}

\paragraph{Annotation and replay masks.}
The paired Fixed UL and SPARK classification runs use sensitivity scores only to partition replay positions at a threshold of 0.6; correction targets use frozen explicit-identifier annotations. ConPriv-Bench aligns registered-email annotations with tokenization to obtain binary target masks. On both benchmarks, SPARK's current-task anchor covers valid non-target prompt and response positions. Sensitivity-defined groups in the original classification diagnostics remain distinct from these explicit-identifier targets.

\paragraph{Paired correction configurations.}
Tables~\ref{tab:ptpc_components} and~\ref{tab:simple_controls} report paired complete-task-stream results for Fixed UL and \method{} under shared data and replay schedules. Both use an effective unlikelihood weight $\lambda_{\mathrm{pii}}\alpha_{\mathrm{UL}}=16$ and top-$K$ distillation with $K=100$. Fixed UL uses only unlikelihood correction, whereas \method{} adds the current-task KL anchor specified in Table~\ref{tab:hyperparams}. On ConPriv-Bench, both use 800 correction steps per task with batch size 2; on classification, both use 200 steps with batch size 1. Correction uses a learning rate of $10^{-5}$. Other baselines follow their respective configurations.